\pdfoutput=1

\documentclass[11pt]{article}

\usepackage[]{ACL2023}

\usepackage{times}
\usepackage{latexsym}

\usepackage[T1]{fontenc}

\usepackage[utf8]{inputenc}

\usepackage{microtype}

\usepackage{inconsolata}

\usepackage{amsmath}
\usepackage{amssymb}
\usepackage{xparse}
\usepackage{graphicx}
\usepackage{nicematrix}
\usepackage{booktabs} 
\usepackage[linesnumbered,ruled,vlined,algo2e]{algorithm2e}
\usepackage{caption}
\usepackage{subcaption}

\definecolor{myblue}{rgb}{0,0.44,0.75}
\definecolor{gold}{rgb}{0.83, 0.69, 0.22}
\definecolor{mygreen}{rgb}{0,0.69,0.31}
\definecolor{mypurple}{rgb}{0.44,0.19,0.63}
\newcommand{\red}[1]{{\color{red} #1}}
\newcommand{\datasetname}{OpenPI-C}

\iftrue
\NewDocumentCommand{\xueqing}
{ mO{} }{\textcolor{cyan}{\textsuperscript{\textit{Xueqing}}\textsf{\textbf{\small[#1]}}}}
\NewDocumentCommand{\zoey}
{ mO{} }{\textcolor{orange}{\textsuperscript{\textit{Zoey}}\textsf{\textbf{\small[#1]}}}}
\NewDocumentCommand{\heng}
{ mO{} }{\textcolor{red}{\textsuperscript{\textit{Heng}}\textsf{\textbf{\small[#1]}}}}
\else
\newcommand{\xueqing}[1]{}
\newcommand{\zoey}[1]{}
\newcommand{\heng}[1]{}
\fi

\title{\datasetname: A Better Benchmark and Stronger Baseline for Open-Vocabulary State Tracking}

\author{
	Xueqing Wu$^*$, ~Sha Li$^*$, ~Heng Ji \\
	University of Illinois Urbana-Champaign \\
  \texttt{\{xueqing8,shal2,hengji\}@illinois.edu} \\}

\newcommand\blfootnote[1]{%
  \begingroup
  \renewcommand\thefootnote{}\footnote{#1}%
  \addtocounter{footnote}{-1}%
  \endgroup
}

\begin{document}
\maketitle

\blfootnote{$^*$ Equal contribution}

\begin{abstract}
Open-vocabulary state tracking is a more practical version of state tracking that aims to track state changes of entities throughout a process without restricting the state space and entity space. OpenPI~\cite{tandon-etal-2020-dataset} is to date the only dataset annotated for open-vocabulary state tracking. However, we identify issues with the dataset quality and evaluation metric. For the dataset, we categorize 3 types of problems on the procedure level, step level and state change level respectively, and build a clean dataset OpenPI-C using multiple rounds of human judgment. For the evaluation metric, we propose a cluster-based metric to fix the original metric's preference for repetition.

Model-wise, we enhance the seq2seq generation baseline by reinstating two key properties for state tracking: temporal dependency and entity awareness. The state of the world after an action is inherently dependent on the previous state. We model this dependency through a dynamic memory bank and allow the model to attend to the memory slots during decoding. On the other hand, the state of the world is naturally a union of the states of involved entities. Since the entities are unknown in the open-vocabulary setting, we propose a two-stage model that refines the state change prediction conditioned on entities predicted from the first stage. Empirical results show the effectiveness of our proposed model especially on the cluster-based metric. The code and data are released at \url{https://github.com/shirley-wu/openpi-c}

\end{abstract}

\section{Introduction}

\begin{figure*}[!htb]
	\centering
	\includegraphics[width=\linewidth]{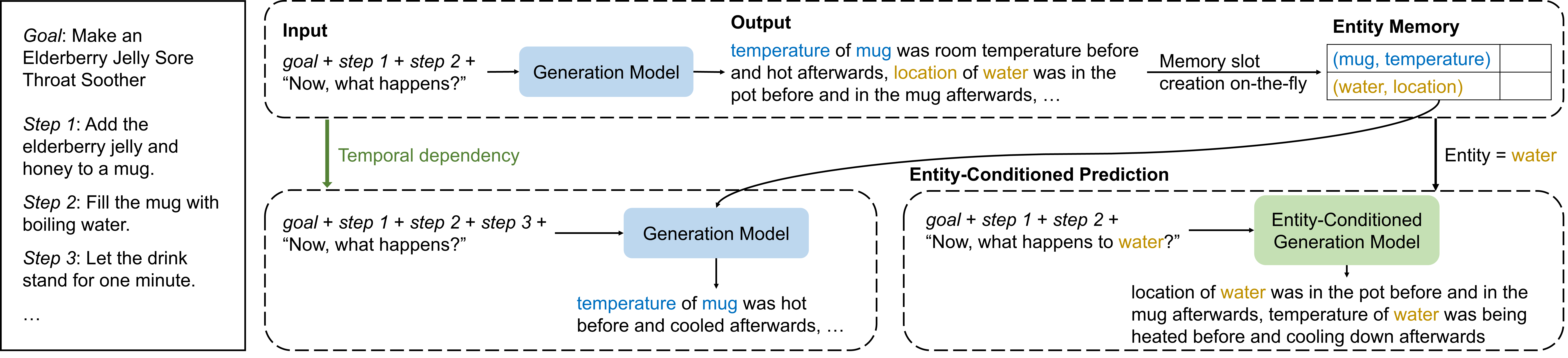}
	\caption{
 The baseline generation model for open-vocabulary state tracking takes the goal and previous steps as input and generates the state changes as templated sentences. We propose to model temporal dependency between steps using an entity memory module and increase entity awareness of the model by using a two-stage procedure where the second state prediction is conditioned on entities from the first stage.
 }
	\label{fig:overall}
\end{figure*}

State tracking is the task of predicting the states of the world after an action is performed.
Most existing work operate under a simplified \textbf{close-vocabulary} setting, assuming the state space and involved entities are known~\cite{dalvi-etal-2018-tracking, DBLP:conf/iclr/BosselutLHEFC18}, which limits their applicability.
The more practical \textbf{open-vocabulary} setting assumes both the entities and the state space are unknown. The OpenPI dataset~\citep{tandon-etal-2020-dataset} is, to our knowledge, the first and only dataset for this task. However, we find a series of issues concerning data quality and evaluation, which may hinder progress in this line of research. 

We identify three types of issues with the dataset: non-procedural documents, out-of-order steps, and ambiguous state changes. 
In particular, $\sim$32\% of the state changes cannot be reliably inferred from the input, which we find encourages model hallucination. 
We filter out problematic data points 
and build a cleaner dataset via crowdsourcing. 

For evaluation, the greedy matching strategy employed by \citet{tandon-etal-2020-dataset} allows matching multiple predicted state changes to a single gold state change, inadvertently inflating the score when the model produces repetitive outputs.
We propose a \textbf{cluster-based metric} that automatically merges repetitive stage changes and enforces 1-to-1 assignment between clusters. 

We propose two enhancements to the seq2seq generation model proposed for this task in \citet{tandon-etal-2020-dataset}.
To capture the dependency between world states of consecutive time steps, we introduce an \textbf{entity memory} to preserve information about the world state for all previous steps.
When predicting the state changes for subsequent actions, the model can access the state information of previous time steps.
Additionally, while close-vocabulary setting usually provides a list of involved entities to track, such a list is inaccessible in open-vocabulary setting. This requires the model to jointly identify involved entities and predict their state changes.
To make the problem more tractable and help model learning, we propose an \textbf{entity-conditioned prediction step} where predictions are conditioned on each single entity extracted from the predictions of the first stage. 

Our contributions can be summarized as follows:
(1) we present a clean dataset OpenPI-C for open-vocabulary state tracking which fixes the data quality issues in the original OpenPI dataset;
(2) we design a clustering-based metric for state tracking evaluation that mitigates the original metric's preference for repetition;
(3) we model temporal dependency and entity awareness by enhancing the generation model for open-vocabulary state tracking with a dynamic memory module and two-stage prediction. 

\section{Related Work}
Most existing work on entity state tracking~\cite{Weston2016babi, dalvi-etal-2018-tracking, DBLP:conf/iclr/BosselutLHEFC18} is closed-vocabulary, assuming that the number of possible states and involved entities is limited and known.
Under this setting, state tracking can be modeled as a tagging problem~\cite{gupta-durrett-2019-tracking,amini2020procedural, huang-etal-2021-reasoning} which is not applicable for the open-vocabulary case.
\citet{tandon-chatterjee-2022-team} proposed OpenPI dataset for the more practical open-vocabulary setting. They formulate the task as a generation problem to handle the open vocabulary challenge.

The design of an external memory component has already been applied to close-vocabulary state tracking~\cite{DBLP:conf/iclr/BosselutLHEFC18,DBLP:conf/emnlp/YagciogluEEI18, gupta-durrett-2019-tracking}. 
However, they rely on known entities and only track a limited set of attributes. 
In this work, we use a dynamic memory that can handle emerging entities with open-vocabulary attributes.
\begin{figure*}[htb]
     \centering
     \begin{subfigure}[b]{0.24\linewidth}
         \centering\includegraphics[width=\linewidth]{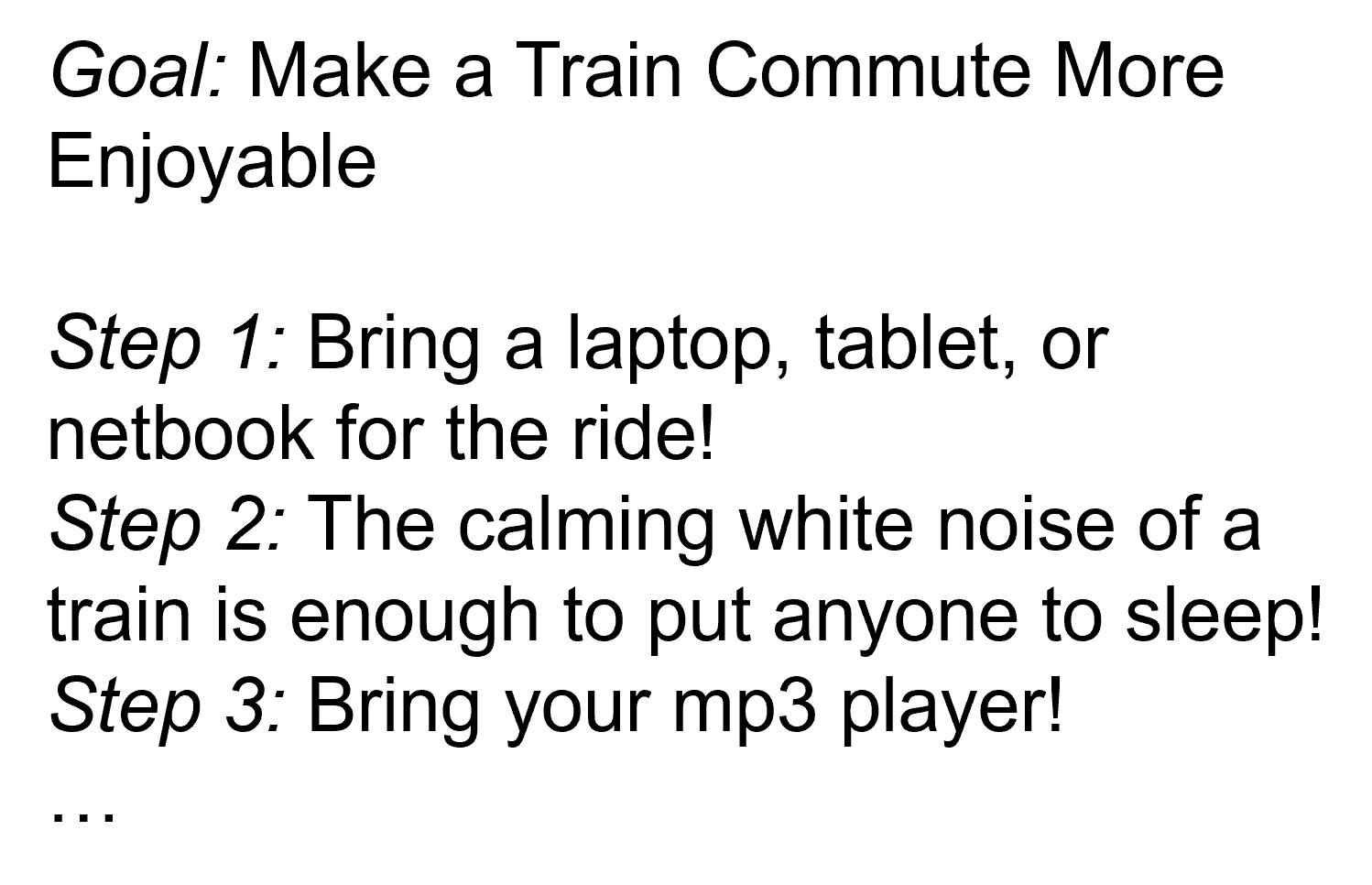}
         \caption{The text is not a procedure text because the steps are not temporally related.}
         \label{fig:examples_filtering:1}
     \end{subfigure}
     \hfill
     \begin{subfigure}[b]{0.4\linewidth}
         \centering
         \includegraphics[width=\linewidth]{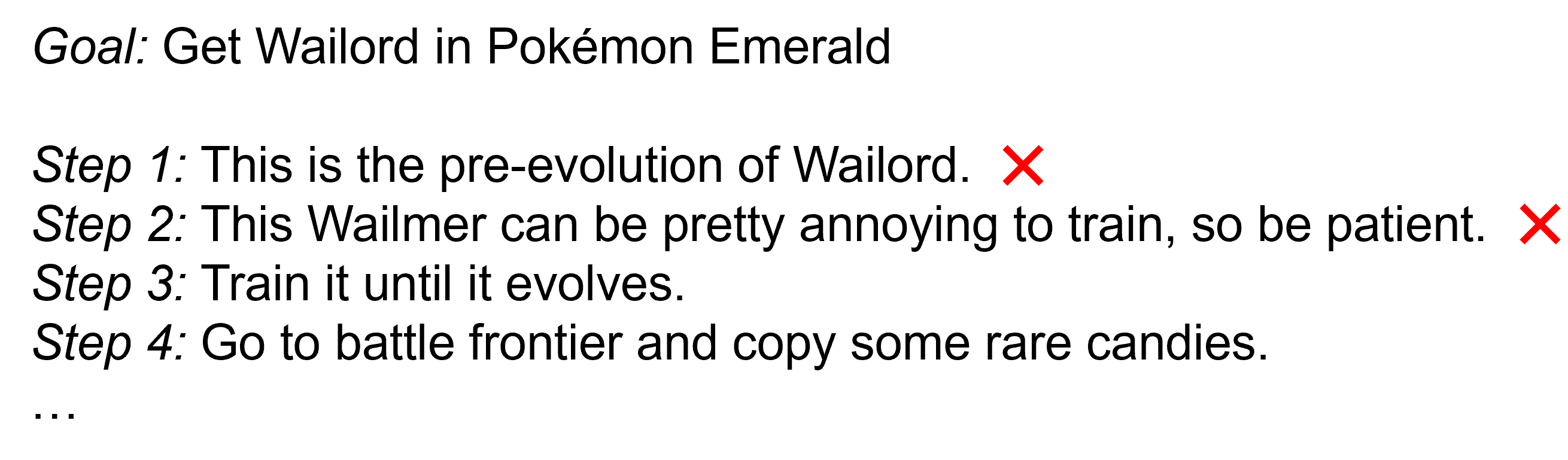}
         \caption{The first and second steps are invalid steps. The first step describes a pre-condition and is not executable. The second step provides complementary information and is not necessary to execute when combined with other steps.}
         \label{fig:examples_filtering:2}
     \end{subfigure}
     \hfill
     \begin{subfigure}[b]{0.3\linewidth}
         \centering
         \includegraphics[width=\linewidth]{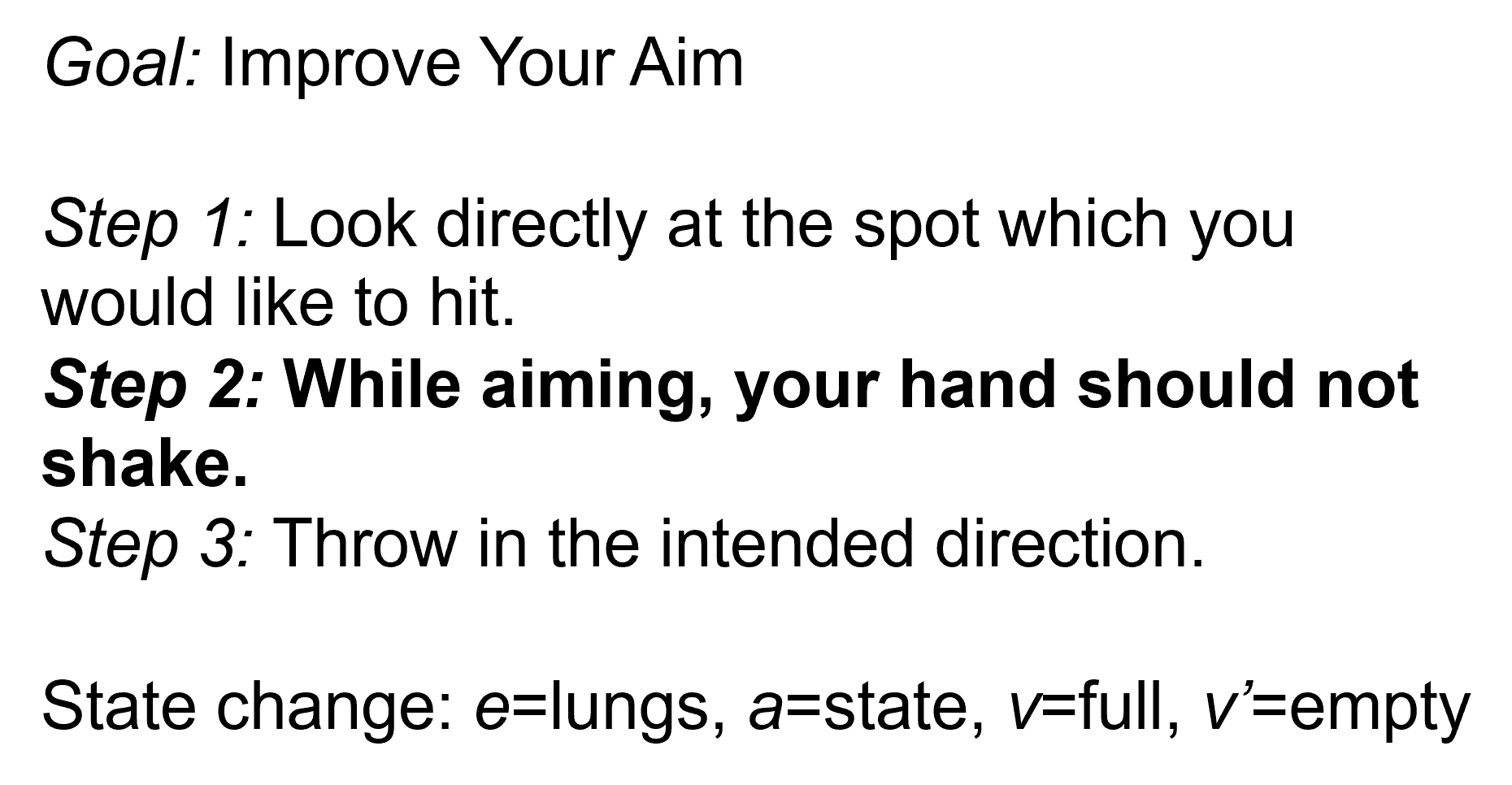}
         \caption{The state change cannot be reliably inferred from step 2. Step 2 involves aiming and stabilizing the hand only, not the lungs.}
         \label{fig:examples_filtering:3}
     \end{subfigure}
        \caption{Examples of low-quality data points removed during the filtering process.}
        \label{fig:examples_filtering}
\end{figure*}
\section{Task and Dataset}

The OpenPI dataset \citep{tandon-etal-2020-dataset} is, to our knowledge, the first and only dataset for open-vocabulary state tracking. The texts are collected from WikiHow and the state changes are manually annotated.

\paragraph{Dataset Issues}
We identify 3 types of quality issues in the OpenPI dataset. For input, we find that $\sim$15\% input texts are not procedure texts because the steps do show any temporal continuity (shown in Figure \ref{fig:examples_filtering:1}). 
In valid procedure text inputs, $\sim$7.4\% steps are invalid steps in the context of the procedure texts (shown in Figure \ref{fig:examples_filtering:2}). They either do not explicitly describe an executable action, or do not follow the temporal order when combined with other steps. For output, $\sim$32\% state changes cannot be reliably inferred from the input (shown in Figure \ref{fig:examples_filtering:3}). Such data will encourage the trained model to generate hallucination.

To address these issues and improve data quality, we build a cleaned dataset named OpenPI-C through three-stage human cleaning: (1) filtering out non-procedure input texts, (2) filtering out invalid steps, and (3) filtering out unreliable state changes. In the three stages, we assign each data point with 3/3/2 annotators respectively and achieve 69.4\%/84.9\%/71.0\% agreement (defined as the ratio of data points where all annotators agree with each other). To verify the annotation quality, we manually annotate 50 instances for each stage. 90\%/92\%/84\% of the crowd-sourcing annotations match our manual annotations for the three stages respectively. The statistics of the original OpenPI dataset and our OpenPI-C dataset are presented in Table \ref{tab:statistics}. Detailed annotation settings and filtering criteria are in Appendix \ref{appendix:data_filtering}. Though our dataset has fewer data samples, as shown in Figure \ref{fig:examples_filtering}, the removed data samples are mostly of low quality. As shown in Figure \ref{fig:data_cleaning_model_output}, including such samples in the dataset encourages hallucination and negatively impacts model performance. 

\begin{table}[!htb]
    \centering
    \setlength{\tabcolsep}{3.5pt}
    \small
    \begin{NiceTabular}{l|rrr|rrr}
        \toprule
        & \Block{1-3}{\textbf{OpenPI}} & & & \Block{1-3}{\textbf{OpenPI-C}} & & \\
        & Train & Dev & Test & Train & Dev & Test \\
        \midrule
        \# procedure texts & 644 & 55 & 111 & 539 & 50 & 74 \\
        \# steps & 3216 & 274 & 560 & 2403 & 219 & 345 \\
        \# state changes & 23.9k & 1.7k & 4.2k & 13.8k & 1.2k & 2.0k \\
        \bottomrule 
    \end{NiceTabular}
    \caption{Statistics of the original OpenPI dataset and our OpenPI-C dataset.}
    \label{tab:statistics}
\end{table}

\paragraph{Evaluation Issues}\label{sec:evaluation}
In \citet{tandon-etal-2020-dataset}, each predicted state is matched to the ground truth state with the highest similarity.
As a result, when the model generates near-duplicate state changes, it will artificially boost the model's score. 
We propose a \textbf{cluster-based metric} to address this issue.
We cluster the predicted set and the gold-standard set respectively based on Sentence-BERT \citep{reimers-gurevych-2019-sentence} embedding similarity. After obtaining the predicted and gold-standard clusters, we assign a gold-standard cluster for each predicted cluster through maximal matching which enforces one-to-one mapping. 
Eventually, we use the assignment to calculate precision, recall and F1 scores.




\section{Method}

\paragraph{Generation Baseline}
As shown in Figure \ref{fig:overall}, the input to the model is the concatenation of the goal, steps, and a prompt ``\textit{Now, what happens?}''.
In \citet{tandon-etal-2020-dataset}, each state change will be represented as a templated sequence for generation.
For example, \textit{(potato, shape, whole, cut in half)} will be converted to \textit{``shape of potato was whole before and cut in half afterwards''}. 

\paragraph{Entity Memory}
To capture the temporal dependency across steps, we maintain a variable-size memory bank to store historical state changes.
For each entity-attribute pair $(e, a)$ that appears in the prediction, we allocate a memory slot after it first appears in the predicted state changes. Suppose it first appears at step $k_0$, then we initialize its memory $\mathbf{m}$ at the next step $\mathbf{m}^{k_0 +1} = \mathbf{h}^{k_0}$. Here, $\mathbf{h}^{k_0}$ represents the hidden states for $(e,a)$ at step $k_0$.
In the subsequent steps, we update the memory every time the attribute $a$ of entity $e$ changes. Formally, at step $k, k>k_0$, if $(e,a)$ changes, then $\mathbf{m}^{k+1} = \left(\mathbf{m}^{k} + \mathbf{h}^{k}\right) / 2$; otherwise, $\mathbf{m}^{k+1} = \mathbf{m}^{k}$.
To compute $\mathbf{h}^{k}$, we take the text expressing its state change from the 
generated sequence at step $k$ and compress their decoder-side hidden states $\mathbf{h}_1, \ldots, \mathbf{h}_{n}$ into $\mathbf{h}^{k}$ via attention:
\begin{align}
    \alpha_i = \underset{i}{\text{softmax}} \left(\mathbf{W}^{k-k_0} \mathbf{h}_i\right), \mathbf{h}^{k} = \sum_{i=1}^n \alpha_i \mathbf{h}_i
\end{align}
where $\mathbf{W}^{k-k_0}$ is a learnable parameter for the $(k-k_0)$-th step after $(e,a)$ appears. To reduce the number of parameters, we share the same $\mathbf{W}^{k-k_0}$ among all $k,k-k_0>0$. That is, we use $\mathbf{W}^{0}$ to initialize the memory when $(e,a)$ first appears, and use another parameter $\mathbf{W}^{>0}$ to update the memory.

We incorporate the memory through the decoder side cross-attention.
At step $k$, the keys and values for the cross-attention module include two parts: the encoder-side hidden states $\mathbf{h}^{enc}_1\ldots\mathbf{h}^{enc}_n$ ($n$ refers to the number of tokens encoded by the encoder) and the memory vectors $\mathbf{m}^k_1\ldots\mathbf{m}^k_M$ ($M$ refers to the number of created memory slots). We project them into key and value matrices $K, V$ with different parameters:
\begin{align}
\hspace{-1pt} \{K,V\} = & [ \mathbf{W}^{enc}_{\{K,V\}} \mathbf{h}^{enc}_1, \ldots, \mathbf{W}^{enc}_{\{K,V\}} \mathbf{h}^{enc}_n, \\
& \mathbf{W}^m_{\{K,V\}} \mathbf{m}^k_1, \ldots, \mathbf{W}^m_{\{K,V\}} \mathbf{m}^k_M ]^\top, \nonumber
\end{align}
and feed them into the cross-attention module.
In this way, the model can adaptively select between input information and historical state change information stored in the memory.

\paragraph{Entity-Conditioned Prediction}

A challenge for this open-vocabulary task is the lack of access to the entities involved.
Compared to directly modeling all state changes $p(Y|x, g)$ given the steps $x$ and goal $g$, we can decompose this problem into first predicting entities, and then modeling the state change of each entity separately $p(Y_e| x, g, e)$. Conditioning on the entity simplifies the task and eases model training.

We reuse the baseline model and replace the natural language prompt with ``\textit{Now, what happens to $e$?}''.
During inference, 
we extract all the entities in the prediction and perform entity-conditioned prediction for each entity $e$.
Eventually we merge the $N$ sets of state changes as the final output.

\begin{table}[!htb]
    \centering
    \setlength{\tabcolsep}{1.5pt}
    \small
    \begin{NiceTabular}{l|cccccc}
        \toprule
        & \Block{1-3}{\textbf{F1 original}} & & & \Block{1-3}{\textbf{F1 cluster-based}} \\
        & {\fontsize{8pt}{8pt}\selectfont Exact} & {\fontsize{8pt}{8pt}\selectfont BLEU} & {\fontsize{8pt}{8pt}\selectfont ROUGE} & {\fontsize{8pt}{8pt}\selectfont Exact} & {\fontsize{8pt}{8pt}\selectfont BLEU} & {\fontsize{8pt}{8pt}\selectfont ROUGE} \\
        \midrule
        GPT-2 & 3.92 & 20.81 & 39.73 & 5.72 & 20.31 & 33.40 \\
        BART & 4.88 & 23.35 & 41.88 & 7.10 & 22.72 & 35.44 \\
        \ +concat states & 4.73 & 21.96 & 40.38 & 6.69 & 20.61 & 32.88 \\
        \midrule 
        BART+EMem & 5.27 & 24.06 & \textbf{42.71} & 7.65 & 23.40 & \textbf{35.79} \\
        \ +ECond & \textbf{5.70} & \textbf{23.81} & 42.14 & \textbf{8.27} & \textbf{23.56} & \textbf{35.80} \\
        \ +EMem+ECond & 5.65 & 23.73 & 42.15 & \textbf{8.26} & 22.96 & 35.34 \\
        \bottomrule 
    \end{NiceTabular}
    \caption{Main results on \datasetname~(in \%). EMem denotes Entity Memory and ECond denotes Entity-Conditioned prediction.}
    \label{tab:main_results}
\end{table}

\begin{figure}[!htb]
	\centering
	\includegraphics[width=\linewidth]{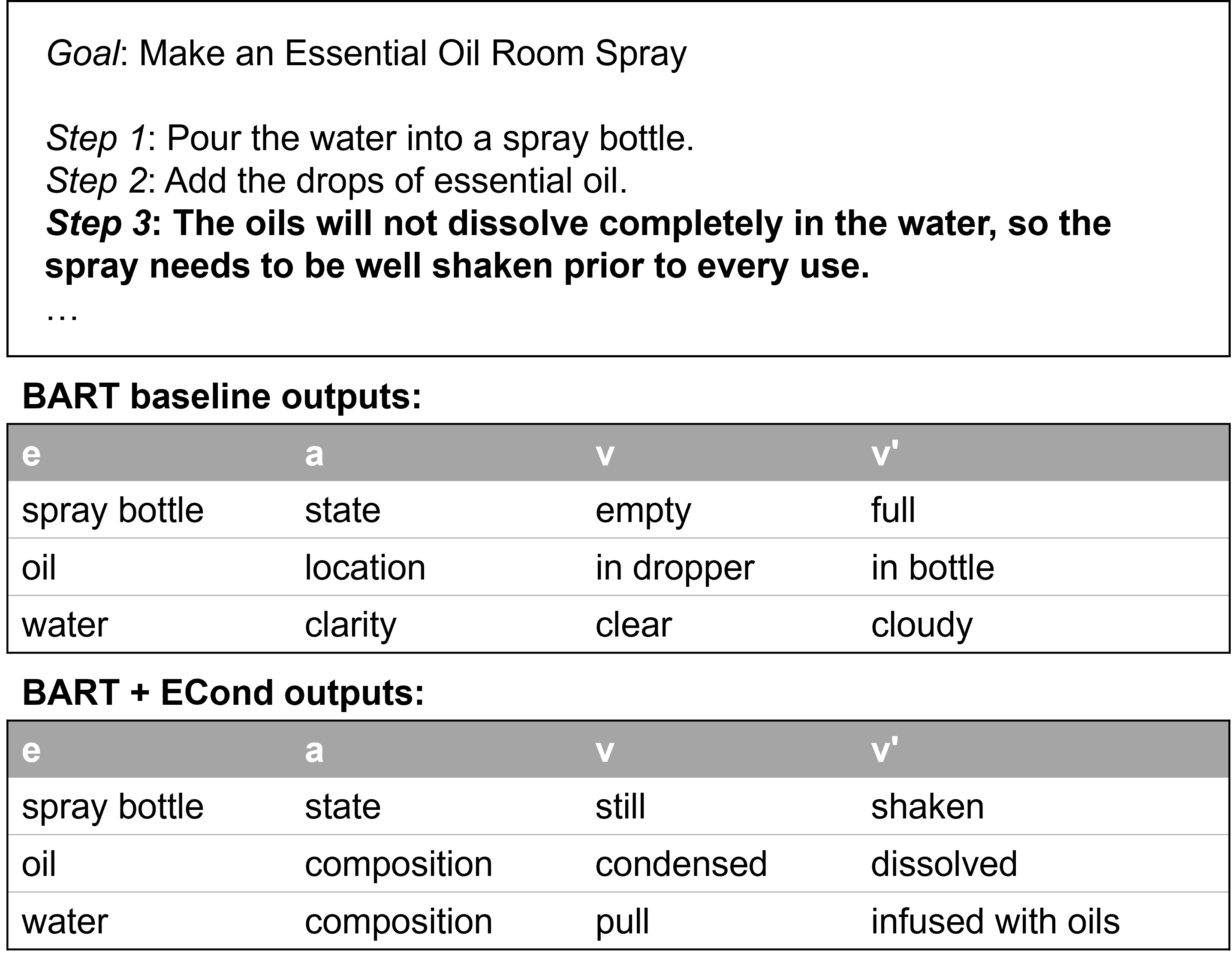}
    \caption{A good case of entity-conditioned prediction (ECond). Based on the same set of entities, entity-conditioned prediction is able to correct the prediction for entities \texttt{spray bottle} and \texttt{oil} and choose more appropriate wording for \texttt{water}.}
    \label{fig:case_study_entcond}
\end{figure}

\section{Experiments}

Our experiments are based on pre-trained BART \citep{DBLP:conf/acl/LewisLGGMLSZ20}.\footnote{Our proposed techniques can be applied on any encoder-decoder model. Among the base models that we have experimented with, we found BART to work the best and hence our experiments are based on BART.}
We add another baseline that that concatenates all previous state changes to the input (denoted as ``BART + concat states'').
Following \citet{tandon-etal-2020-dataset}, we also use GPT-2 \citep{radford2019language} as baseline.

\begin{figure}[!htb]
	\centering
	\includegraphics[width=\linewidth]{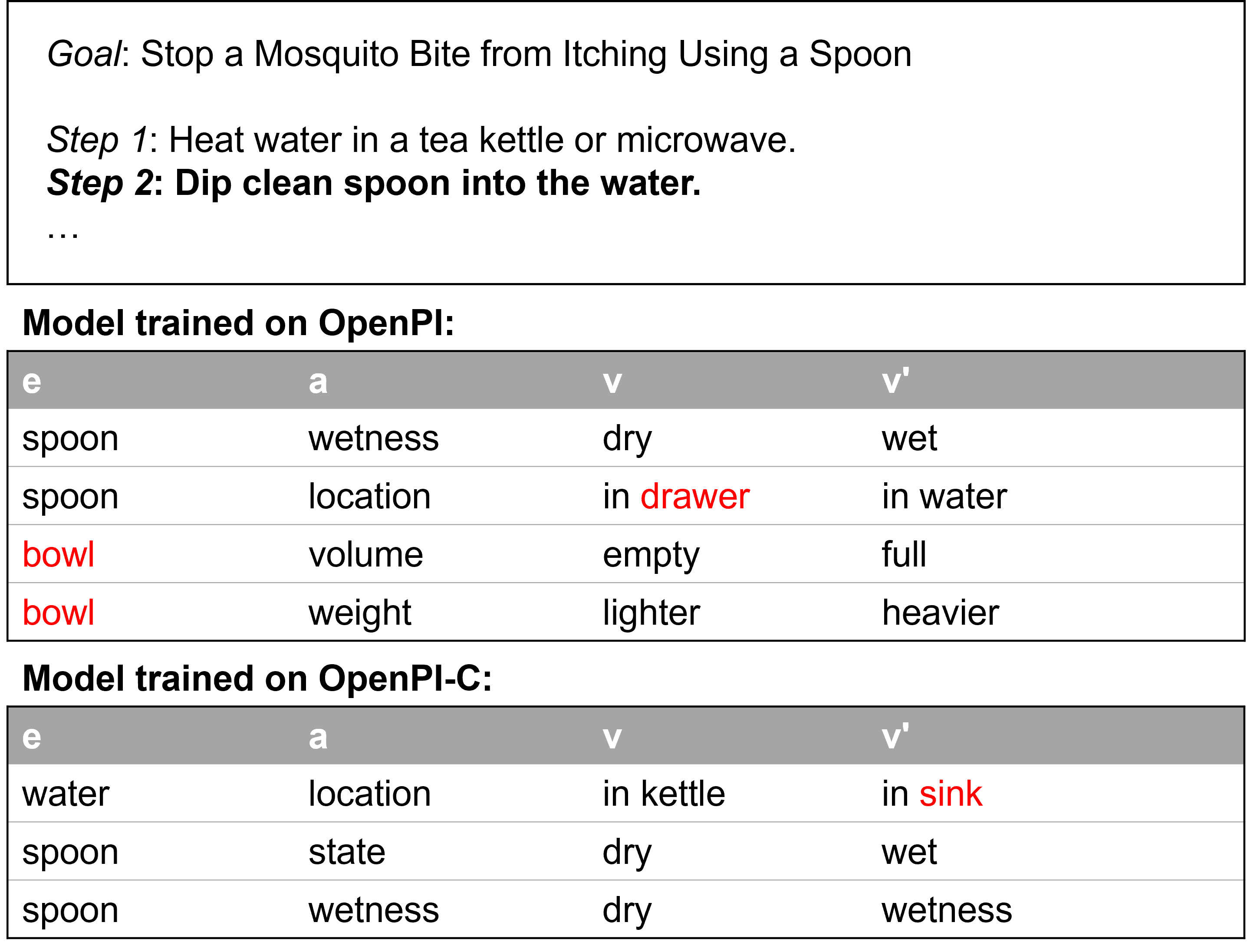}
    \caption{Outputs of our model (BART+EMem+ECond) trained on OpenPI and OpenPI-C  respectively. The model trained on OpenPI produces more hallucination (highlighted in \red{red}).}
    \label{fig:data_cleaning_model_output}
\end{figure}

The main results are in Table \ref{tab:main_results}.
Overall, our proposed two techniques improve performance on most metrics especially on the cluster-based metrics.
Compared to our proposed entity memory (EMem), ``BART + concat states'' takes the same information (historical steps and historical state changes) as input but significantly decreases the performance compared to the baseline.
This is due to the historical state changes being too long and distracting the model.
As in Figure \ref{fig:case_study_entcond}, entity-conditioned prediction (ECond) is able to produce more accurate outputs based on the same set of entities.
We observe that performance gains brought by entity-conditioned prediction are more significant on cluster-based F1 metrics, because the baseline model produces longer and more repetitive outputs (average number of output state changes per step is $7.71$ compared to $6.76$ of BART+ECond). As a result, the original F1 gives the baseline too much credit.

To analyze the effect of dataset cleaning, we compare the outputs of models trained on the original dataset and cleaned dataset. As in Figure \ref{fig:data_cleaning_model_output}, the cleaned dataset encourages the model to stick to the input text and produce less hallucination. To quantify this effect, we manually examined 50 processes randomly sampled from the test set. Of the 50 processes we examined, each process consists of multiple steps, and each step has multiple output state changes. We did a binary classification on each output state change to classify whether it contains hallucinations or not. Overall, the model trained on OpenPI produced 749 hallucinated state changes while the model trained on OpenPI-C produced 393 (47.53\% less).

\section{Conclusion and Future Work}

In this paper we study the open vocabulary state tracking problem.
We build upon the generation formulation introduced by \citet{tandon-etal-2020-dataset} and propose two techniques: (1) \textit{entity memory} that models the temporal dependency by storing world states from previous steps, and (2) \textit{entity-conditioned prediction} that simplifies the task by predicting state changes conditioned on each single entity. We conduct human annotation to address data quality issues in the existing OpenPI dataset and thus propose a cleaned version of OpenPI dataset.
We propose an improved cluster-based metric to overcome the original metric's preference towards repetition.
For future work, we consider using external resources such as ConceptNet \citep{DBLP:journals/ir/AmigoGAV09} to assist entity prediction. 
\section{Limitations}
The scope of this work is limited by the available data.
The OpenPI dataset \citep{tandon-etal-2020-dataset} is derived from WikiHow~\footnote{\url{https://www.wikihow.com/Main-Page}}, and focuses on everyday scenarios and contains English only. 
We would like to see resources that span more domains (e.g. scientific domains) and more languages.

\section{Ethical Considerations}
Our work does not involve the creation of new datasets. However, we would like to point out that the existing dataset OpenPI is based on WikiHow, 
which is primary crowdsourced (with partial expert review). Thus some of the content is influenced by the cultural and educational background of the annotators. In our human cleaning, we recruit annotators from United States and Canada regions only, which may also bring cultural bias to the content.
In particular, some procedures are related to healthcare and neither the procedure nor the model output should be regarded as medical advice.

\section*{Acknowledgement}
This research is based upon work supported by U.S. DARPA DARPA KAIROS Program No. FA8750-19-2-1004. The views and conclusions contained herein are those of the authors and should not be interpreted as necessarily representing the official policies, either expressed or implied, of DARPA, or the U.S. Government. The U.S. Government is authorized to reproduce and distribute reprints for governmental purposes notwithstanding any copyright annotation therein.

\bibliography{anthology,custom}
\bibliographystyle{acl_natbib}

\appendix

\section{Clustering Algorithm}

For output clustering, we use \texttt{stsb-distilroberta-base-v2} model provided by \texttt{sentence-transformers} package\footnote{\url{https://github.com/UKPLab/sentence-transformers}} to obtain sentence embeddings. We use cosine similarity to compute similarity scores.
The detailed algorithm is in Algorithm \ref{alg:clustering}. We set the threshold $th$ as $0.7$.
To evaluate the performance, we manually cluster the outputs for 20 processes (containing 85 steps) and use the annotated clusters as gold clusters to evaluate our algorithm. We calculate BCubed metrics \citep{DBLP:journals/ir/AmigoGAV09} and our algorithm achieves 88.00\% precision, 88.68\% recall, and 87.39\% F1.

\newcommand\mycommfont[1]{\textcolor{blue}{#1}}
\SetCommentSty{mycommfont}
\SetFuncSty{text} 
\SetKwInput{KwInput}{Input}
\SetKwInput{KwOutput}{Output}
\SetKw{Break}{break}

\begin{algorithm2e}[!htb]
\DontPrintSemicolon
  
  \SetKwBlock{RepeatForever}{repeat}{}
  \SetKw{Break}{break}
  
  \KwInput{input set $\mathbf{y}=\{y_1,\ldots,y_n\}$; similarity scorer $S(\cdot,\cdot)$; threshold $th$}
  \KwOutput{clusters $\mathcal{C} = \{C_1,\ldots,C_K\}$}
    
    $\mathcal{C} \gets []$\;
    
    \For{$i \gets 1$ \KwTo $n$}{
      new\_cluster $\gets true$ \;
      
      \For{$k \gets 1$ \KwTo $|\mathcal{C}|$}{
        \uIf{\upshape $\forall y \in C_j, S(y_i, y) > th$}{
          \tcc{Assign $y_i$ to cluster $C_k$}
          
          $C_k$.add($y_i$) \;
          
          new\_cluster $\gets false$ \;
          
          \Break
        }
      }
      
      \uIf{\upshape new\_cluster }{
        $\mathcal{C}$.append($\{y_i\}$)
      }
    }
\caption{The clustering algorithm. The input set $\mathbf{y}$ is the gold or the predicted set of state changes. Each output cluster $C_k$ is a subset of $\mathbf{y}$ and all output clusters $\mathcal{C}$ form a partition of $\mathbf{y}$.}
\label{alg:clustering}
\end{algorithm2e}

\section{Data Details}
\label{appendix:data_filtering}

OpenPI dataset is released by \citet{dalvi-etal-2018-tracking}.\footnote{The original dataset and their baseline and evaluation code are released at \url{https://github.com/allenai/openpi-dataset}. We notice that some data has wrong template (``$a$ of $e$ were $v$ before and $v'$ afterwards'' instead of ``was''), which will influence model training and evaluation. Thus, we apply a preprocessing step to fix these errors.} It is an English-only dataset crawled from WikiHow and annotated via crowd-sourcing.

As mentioned before, we conduct human annotation to filter out low-quality data. The annotation study is reviewed by an ethics review board and determined to be a not human subjects research. The annotation is conducted on MTurk platform. To ensure that the annotators are native English speakers, we recruit annotators from the United States and Canada. We have informed the annotators in the annotation instructions that we are collecting data for research purpose.
The annotation includes three stages:

\textbf{Stage 1}: filter out non-procedure texts. Each annotator is presented with an input text and asked to judge whether it is a procedure text or not. Each input text is annotated by three annotators; the reward for annotating each input text is \$0.03. We remove input texts that are considered as non-procedure texts by most annotators (i.e., at least two annotators). 15\% procedure texts are removed at this stage.

\textbf{Stage 2}: filter out invalid steps. Each annotator is presented with an procedure text. For each step in the process, the annotator is asked to judge whether it is a valid step. Each input text is annotated by three annotators; the reward for annotating each input text is \$0.2. We then remove steps that are considered as invalid steps by most annotators (i.e., at least two annotators). 7.4\% steps are removed at this stage.

\textbf{Stage 3}: filter out low-quality state changes.
Each annotator is presented with an input procedure text and a state change caused by one of the steps. The annotator is asked to decide whether the state change is \textit{certain}, \textit{uncertain} and \textit{impossible}. Each state change is annotated by two annotators; the reward for annotating each state change is \$0.05. To ensure data quality, we remove state changes that receive at least one \textit{uncertain} or \textit{impossible} rating from the two annotators, which empirically yield the best results. 32\% state changes are removed at this stage.

\begin{figure*}[htb]
     \centering
     \begin{subfigure}[b]{0.31\linewidth}
         \centering\includegraphics[width=\linewidth]{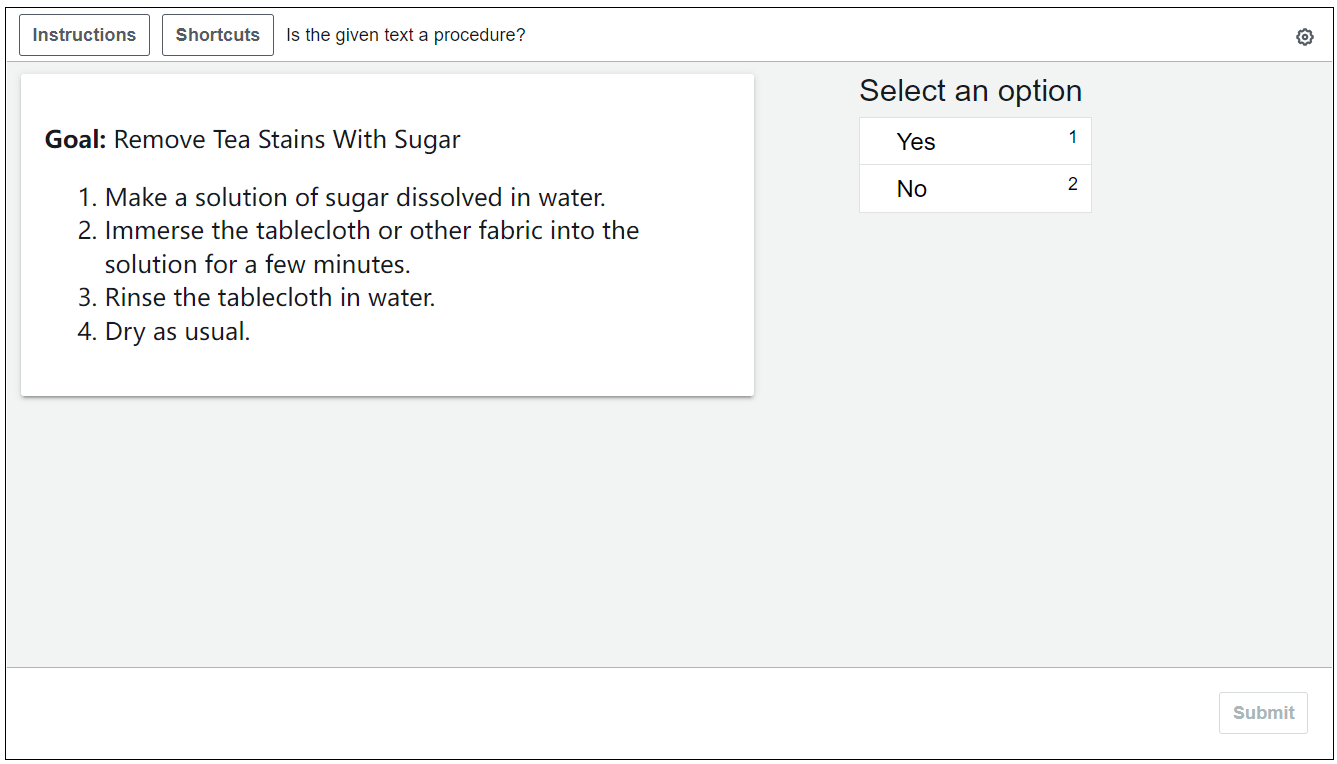}
         \caption{Stage 1.}
     \end{subfigure}
     \hfill
     \begin{subfigure}[b]{0.31\linewidth}
         \centering
         \includegraphics[width=\linewidth]{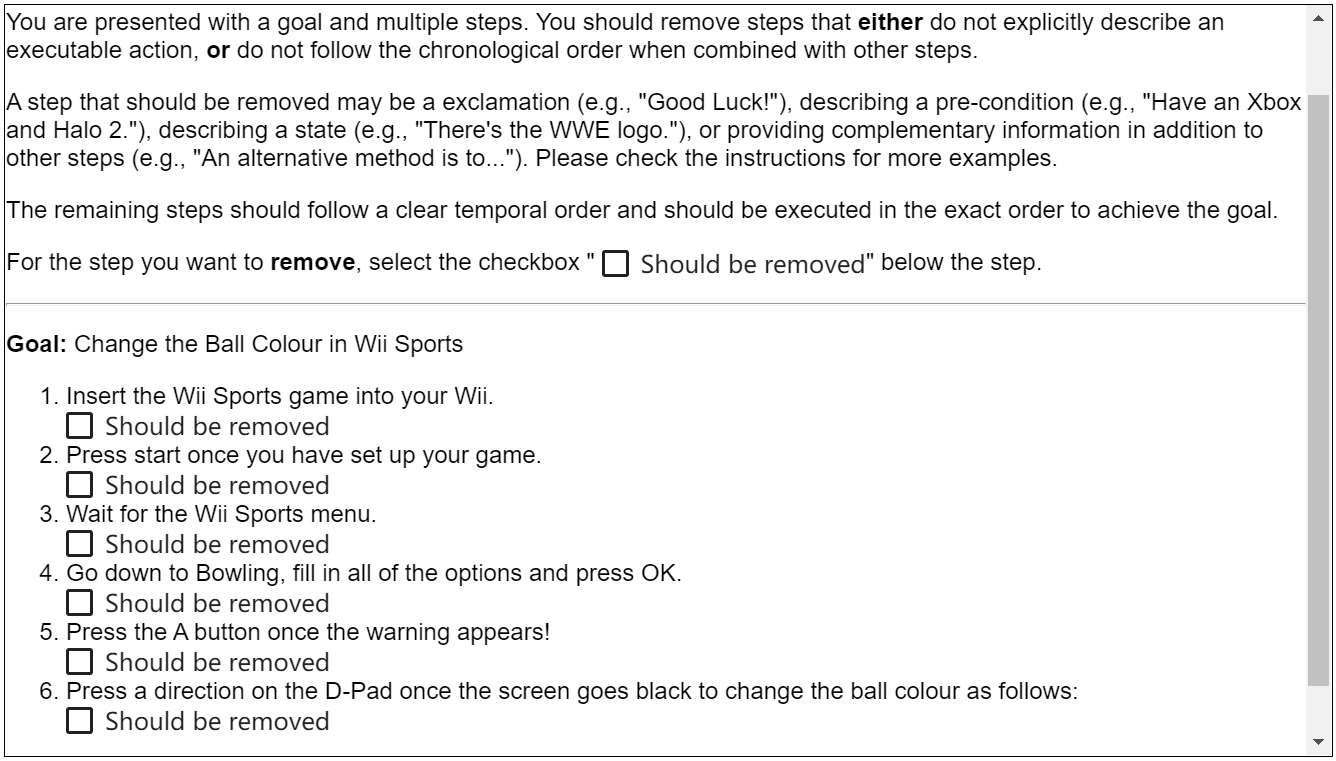}
         \caption{Stage 2.}
     \end{subfigure}
     \hfill
     \begin{subfigure}[b]{0.33\linewidth}
         \centering
         \includegraphics[width=\linewidth]{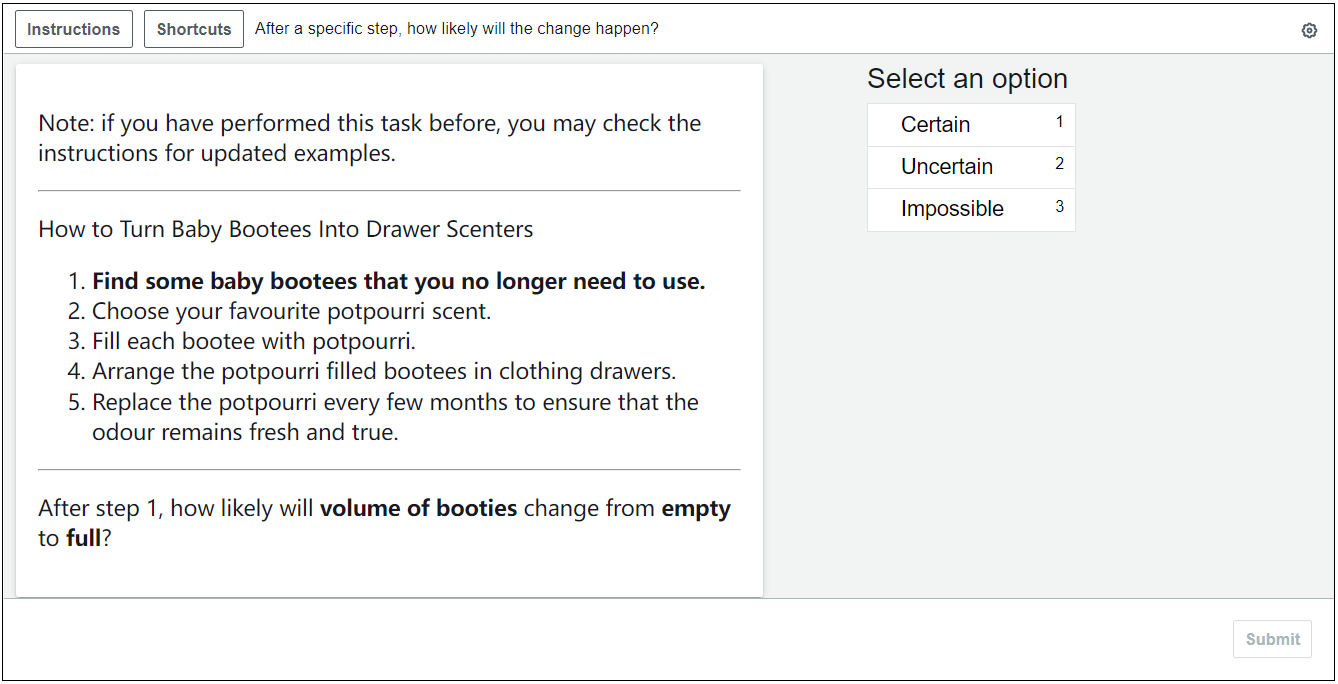}
         \caption{Stage 3.}
     \end{subfigure}
        \caption{Screenshots of the annotation interface.}
        \label{fig:annotation}
\end{figure*}

\begin{table*}[!htb]
    \centering
    \begin{NiceTabular}{l|cccccc}
        \toprule
        & \Block{1-3}{\textbf{F1 original}} & & & \Block{1-3}{\textbf{F1 cluster-based}} \\
        & Exact & BLEU & ROUGE & Exact & BLEU & ROUGE \\
        \hline
        GPT-2 \citep{tandon-etal-2020-dataset} & 4.3 & 16.1 & 32.4 & - & - & - \\
        GPT-2 & 5.35 & 19.57 & 36.26 & 6.16 & 18.24 & 29.95 \\
        BART & 5.51 & 23.19 & 40.45 & 7.40 & 21.44 & 33.22 \\
        \ +concat states & 4.65 & 19.58 & 36.82 & 6.22 & 18.20 & 29.99 \\
        \midrule 
        BART+EMem & 6.15 & 23.63 & \textbf{40.60} & 7.81 & 21.69 & \textbf{33.30} \\
        \ +ECond & 6.88 & 23.50 & 40.22 & 9.16 & 22.29 & 33.15 \\
        \ +EMem+ECond & \textbf{7.38} & \textbf{23.71} & 40.33 & \textbf{9.69} & \textbf{22.38} & 33.02 \\
        \bottomrule 
    \end{NiceTabular}
    \caption{Results (in \%) on the original OpenPI dataset. EMem denotes Entity Memory and ECond denotes Entity-Conditioned prediction.}
    \label{tab:main_results_old}
\end{table*}

\begin{table*}[!htb]
    \centering
    \begin{NiceTabular}{l |cc|cc|c }
        \toprule 
        & \Block{1-2}{\textbf{Top-p sampling}} & & \Block{1-2}{\textbf{Top-k sampling}} & & \textbf{Beam search} \\
        & \ p=0.9\ & \ p=0.5\ & \ \ \ k=10\ \ \ & \ \ \ k=5\ \ \ & beam=4 \\
        \midrule 
        GPT-2 & 16.78 & 17.38 & 16.49 & 17.29 & 18.24 \\
        BART & 19.94 & 20.31 & 19.96 & 19.45 & 21.44 \\
        Ours & 19.64 & 21.65 & 20.68 & 20.45 & 22.38 \\
        \bottomrule 
    \end{NiceTabular}
    \caption{Results (in \%) of GPT-2 baseline, BART baseline, and our proposed method with different decoding strategies on the original OpenPI dataset. We report clustering-based F1 with BLEU. Among all settings, beam search achieves the best performance.}
    \label{tab:decoding_strategy}
\end{table*}

Screenshots of the annotation interface are shown in Figure \ref{fig:annotation}.
Eventually, we manually examine the data and conducted rule-based filtering according to the following heuristics. We first remove steps with no state changes, and then remove procedure texts with $<3$ steps.

\section{Experiment Details}

We use GPT2-medium and BART-large models for the experiments.
The number of parameters for GPT-2 baseline, BART baseline, BART+EMem and BART+ECond models are $355$M, $406$M, $444$M and $406$M respectively.
Each experiment is run on one Telsa P100 GPU and takes about 4 hours. 

In training, we use the exact training hyperparameters as \citet{dalvi-etal-2018-tracking}, i.e., the learning rate of $5\times 10^{-5}$, the batch size of $8$, and $30$ epochs.

\begin{figure}[!htb]
	\centering
	\includegraphics[width=\linewidth]{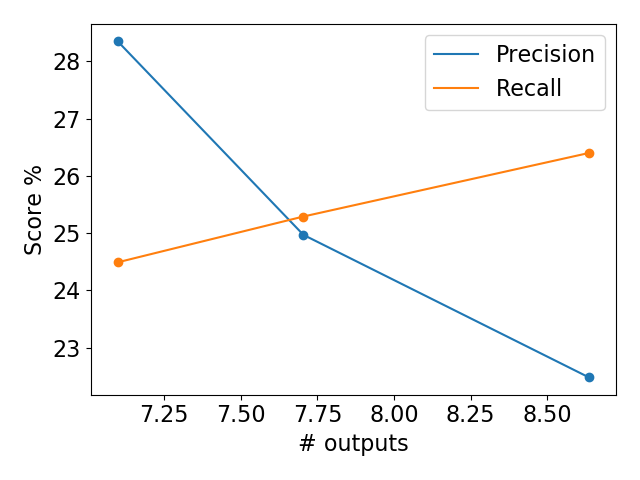}
    \caption{Precision and recall under different numbers of outputs for BART baseline. The length penalty is set as $0.2$, $1.0$ and $2.0$ respectively.}
    \label{fig:precision_recall}
\end{figure}

In decoding, we use beam search with beam size of $4$.
The decoding strategy is searched from top-p sampling ($0.5\le p \le 0.9$), top-k sampling ($5\le k \le 10$) and beam search (beam$=4$).
The best decoding strategy is found by manual tuning on the original OpenPI dataset. Results are in Table \ref{tab:decoding_strategy}. We show that using beam search significantly boost the performance over top-p or top-k sampling for all systems.
We also show in Figure \ref{fig:precision_recall} that length penalty can be used to control the number of outputs, and thus to balance between precision and recall.

Compared to \citet{dalvi-etal-2018-tracking}, our re-implemented GPT-2 baseline is different in that: (1) we include the process goal $g$ in the input, and (2) we use beam search with beam size of $4$ instead of top-p sampling.

We also run the experiments on the original OpenPI dataset and compare with the results of \citet{dalvi-etal-2018-tracking}. Results are shown in Table \ref{tab:main_results_old}.

\section{Scientific Artifacts}

Scientific artifacts we use in this work include:
(1) OpenPI dataset \citep{tandon-etal-2020-dataset} and their baseline and evaluation code released under the
MIT License. The dataset is collected from WikiHow and focuses on every-day scenarios and contains English only. Our use is consistent with the resource's intended use, which is to facilitate research on open-vocabulary state tracking tasks. (2) Three pre-trained models: GPT-2 \citep{radford2019language} and BART \citep{DBLP:conf/acl/LewisLGGMLSZ20} provided by \texttt{transformers}\footnote{\url{https://huggingface.co/docs}} and Sentence-BERT \citep{reimers-2019-sentence-bert} provided by \texttt{sentence-transformers}, all licensed
under the Apache License 2.0. 
We use the models for research which is consistent with their intended use. Our code and data are released under the MIT license, which is compatible with the artifacts utilized in our research.

\end{document}